\documentclass[11pt]{article}
\usepackage[utf8]{inputenc}
\usepackage{mathpazo,helvet}
\usepackage{graphicx}
\usepackage{textcomp}
\usepackage{tabularx}
\usepackage{lscape}
\usepackage[round,colon]{natbib}
\usepackage[pdftex,pdfstartview=FitH]{hyperref}

\usepackage{amsmath}
\renewcommand\Vec[1]{\ensuremath{\mathbf{#1}}}
\newcommand\Mat[1]{\ensuremath{\mathbf{#1}}}
\newcommand\SubMat[1]{\ensuremath{\mathrm{#1}}}
\newcommand\trnsp{\ensuremath{\mathsf{T}}}
\usepackage[capitalize,noabbrev]{cleveref}

\title{Joint Deformable Registration of Large EM Image Volumes: A Matrix Solver
Approach}
\author{Khaled Khairy \and Gennady Denisov \and Stephan Saalfeld}

\begin{document}

\maketitle

\begin{abstract}
Large electron microscopy image datasets for connectomics are typically composed
of thousands to millions of partially overlapping two-dimensional images
(tiles), which must be registered into a coherent volume prior to further
analysis.  A common registration strategy is to find matching features between
neighboring and overlapping image pairs, followed by a numerical estimation of
optimal image deformation using a so-called solver program.

Existing solvers are inadequate for large data volumes, and inefficient for
small-scale image registration.

In this work, an efficient and accurate matrix-based solver method is presented.
A linear system is constructed that combines minimization of feature-pair square
distances with explicit constraints in a regularization term. In absence of
reliable priors for regularization, we show how to construct a rigid-model
approximation to use as prior.  The linear system is solved using available
computer programs, whose performance on typical registration tasks we briefly
compare, and to which future scale-up is delegated.  Our method is applied to
the joint alignment of 2.67 million images, with more than 200 million point-pairs
and has been used for successfully aligning the first full adult fruit fly brain.
\end{abstract}

\section{Introduction}

Electron microscopy (EM) images contain anatomical information relevant for
detailed reconstruction of neuronal circuits.  Modern EM acquisition systems
produce increasingly large sample volumes, with millions of images
\citep{ZhengAl2017}.  Due to the limited field-of-view of current imaging
hardware, and the desire to image large volumes, partially overlapping images
(tiles) are acquired, to be registered into a seamless three-dimensional volume
in a post-acquisition processing step.  Analysis is performed on such fully
registered image volumes
\citep{TakemuraAl2015,TakemuraAl2013,BockAl11,SaalfeldAl10,SaalfeldAl12,
SchefferAl13,CardonaAl12}.

A major challenge for EM volume registration is the large dataset size.  For
example, to image a whole fruit-fly brain (approximately 500\,\textmu{}m in its
largest dimension) at voxel dimensions of
4\texttimes4\texttimes40\,nm$^3$ with serial section transmission EM
(ssTEM), with 10\% overlap of adjacent tiles, approximately 21 million images
($\approx$380 tera voxels) are generated \citep{ZhengAl2017}.  There is a need
for robust, efficient and scalable methods for registering such volumes.

This manuscript addresses three stages of EM volume registration:
\begin{description}
  \item[Montage registration]{The process by which partially overlapping images
within the same section are registered to form a seamless mosaic
(\cref{fig:1}b),}
  \item[Rough alignment]{Scaled-down snapshots of whole montages
produced by the previous step are roughly registered to each other across $z$,
and finally},
  \item[Fine alignment]{In which point-matches obtained from step 1,
together with point-matches across sections are used to jointly register the whole volume.}
\end{description}

Each of the above steps is itself a two-step process. First: Matching
point-pairs are found between pairs of images identified as potential neighbors
(\cref{fig:1}a and b (left panel)), based on matching image features
\citep{Lowe04,SaalfeldAl10,SaalfeldAl12}.  Second: Using that set of point
matches, and after deciding on a transformation model, image transformation
parameters are estimated by a solver algorithm.  The algorithm minimizes
the sum of squared distances between all point matches (\cref{fig:1}).  Deforming tiles according
to these estimated transformations ideally results in a seamless registration of
the whole volume (\cref{fig:1}e).

Available solver algorithms for EM use point-match information local to each
tile to find a local solution, move a step towards that solution, iterate
similarly over all tiles and repeat the process until a convergence criterion is
satisfied.  The solver by \citet{Karsh2016}, although scalable to large
problem sizes, is based on the local iterative scheme described above,
regularizes against stage coordinates which are often unreliable, and is not
guaranteed to converge to a global optimum.  The strategy by
\citet{SaalfeldAl10,SaalfeldAl12} is limited to smaller volumes than described
in this work.
Restriction to iterative techniques necessitates additional solver
parameterization (maximum step size, number of iterations and other empirical
termination criteria) and, at the least-squares level, limits accuracy.  Other
techniques, for example that by \citet{TasdizenAl10}, compose pairwise
transformations sequentially and are therefore prone to additive error.

Main contributions of this work:

(a) We provide a formulation of model-based registration in linear algebra
terms, to perform joint optimization of all images with explicit regularization.
We thereby delegate both scalability and convergence of the solver to efficient
established linear algebra computer programs that may or may not be iterative
(and are generally more efficient even when iterative).  For small to moderate
problem sizes, our method thus uses reliable fast direct methods, which are
parameter free at the least-squares level, and outperform all iterative
approaches.  For large problem sizes, known iterative techniques can be
used.  Moreover, nonlinear models, such as higher order polynomials, and
arbitrary spatial dimensions are permitted in our matrix framework.

(b) We show how, in the absence of reliable starting guesses for tile
transformations, a rigid model approximation can be estimated and used for
explicit regularization.  The presented method therefore assumes no prior
knowledge of image orientation or relative scale, but can take full advantage of
priors when provided.

In practice, tiles must be corrected for non-affine lens-distortion prior to
montaging for which we used the method by \citet{KaynigAl10a} followed by
affine normalization.  We further converted the final solution of the method
described in this manuscript to a thin-plate spline \citep{Bookstein89}
warp-field that we applied to the seamlessly stitched montages.  This way,
we retain the overall shape of the deformable registration approximated by an
affine transformation for each tile while avoiding discontinuous seams between
tiles in the montage.

\section{Methods}

\subsection{The least-squares system}

For a system of partially overlapping image tiles (number of tiles =
$n_\mathrm{tiles}$), we assume that we are given a full set of point-matches
between all neighboring tiles.  We set up and solve a regularized
(Tikhonov-like) least-squares system for finding the set of transformations
$\Vec{x}$ that optimally and jointly minimizes the square distances between all
point-matches
\begin{equation}
\label{eq:1}%
\min_\Vec{x} \lVert \Mat{D}( \Mat{A} \Vec{x} - \Vec{b})
\rVert_2^2 + \lambda \lVert \Mat{B} \Vec{x} - \Vec{ d} \rVert^2
\end{equation}
The first term is the primary objective function.  \Mat{A} is a large $m
\times n$ matrix of point-pair correspondences.  $\Vec{x}$ is a column vector
of transformation coefficients of length $n$, and $\Vec{b}$ is a column vector
of length $m$ ($\Mat{A}$ and $\Vec{b}$ are generally sparse).  $n$ is a product
of the total number of transformations  ($n_\mathrm{tiles}$) and the number of
coefficients ($n_\mathrm{c}$) for the type of transformation considered (e.g.
$n_\mathrm{c} = 6$ for two-dimensional affine).  $m$ is the total number of
point-pairs and $\Mat{D}$ is the individual point-match confidence estimate.
The second term is a regularization term in which $\Mat{B}$ is a matrix with
dimensions $n \times n$ and $d$ is a column vector of length $n$ with values
that the solution $\Vec{x}$ is expected to stay close to, i.e. it is an explicit
regularization.  $\Mat{B}$ only has elements on the diagonal. $\lambda$ is a
regularization parameter that controls the relative importance of fidelity of
$\Vec{x}$ to point-match data vs. adherence to values in $\Vec{d}$.  In
practice, we generalize $\lambda$ to be a diagonal matrix which provides the
ability to constrain individual parameters differently if desired.

\subsection{Constructing \Mat{A}: ``No regularization'' case}

Let us first consider solving the first term of \cref{eq:1} without
regularization, i.e. without the second term in \cref{eq:1}.  First, we
construct $\Mat{A}$ and $\Vec{b}$.  Matching point-pairs between overlapping
tile pair $j$ are given by $^j_i{}p$ and $^j_i{}q$, with $j$ indicating
the pair (e.g.: $j = (1,2)$ or $j = (5,8)$) and point index $i$.  For
reference see \cref{fig:1}. $^j_i{}p$ and $^j_i{}q$ are each 2-vectors (with
columns $^j_i{}p_x, ^j_i{}p_y$ and $^j_i{}q_x, ^j_i{}q_y$).  The length of  $^j p$ and $^j q$ is $n_j$, and varies with the number of point-matches
found between tiles in the pair $j$.  $m$ is twice the sum of all $n_j$.

For example, we want to find two-dimensional transformations to register three
overlapping image tiles jointly ($n_\mathrm{tiles} = 3$), where each tile
overlaps with the other two tiles
\begin{equation*}
\tilde{\Mat{A}}\Vec{x} - \tilde{\Vec{b}} =
\begin{bmatrix}
    ^{1,2}\SubMat{P}  & -^{1,2}\SubMat{Q} & 0 \\
    ^{1,3}\SubMat{P}  & 0                 & -^{1,3}\SubMat{Q} \\
    0                 & ^{2,3}\SubMat{P}  & -^{2,3}\SubMat{Q}
\end{bmatrix}
\Vec{x} -
\begin{bmatrix}
    0 \\
    0 \\
    0
\end{bmatrix}
\end{equation*}
where $^j\SubMat{P}$ and $^j\SubMat{Q}$ represent point-match matrix blocks for
two connected tiles.  $^j\SubMat{P}$ lists point coordinates in the first
tile’s coordinate system and $^j\SubMat{Q}$ those in the  second tile’s.  The
structure of these blocks depends on the transformation model chosen and
reflects the basis functions considered.

For example, in case of an affine model, each point $(x,y)$ in tile coordinates
is transformed to the common coordinate system $(u,v)$ such that
\begin{align*}
u &= a_1 x + a_2 y + a_0 \\
v &= a_4 x + a_5 y + a_3
\end{align*}

The parameters $a_0, a_1, \dots, a_6$ are the transformation parameters sought.
 The number of columns of \Mat{A} is $n = (n_\mathrm{parameters}) \times
(n_\mathrm{tiles}) = 18$, and the point-match blocks are given by
\scriptsize%
\begin{equation*}
^j\SubMat{P} =
\begin{bmatrix}
^j_1p_x     & ^j_1p_y     & 1       & \dots & & \\
\vdots      & \vdots      & \vdots  & \dots & & \\
^j_{n_j}p_x & ^j_{n_j}p_y & 1       & \dots & & \\
            &             &         & ^j_1p_x     & ^j_1p_y     & 1      \\
            &             &         & \vdots      & \vdots      & \vdots \\
            &             &         & ^j_{n_j}p_x & ^j_{n_j}p_y & 1
\end{bmatrix},~
^j\SubMat{Q} =
\begin{bmatrix}
^j_1q_x     & ^j_1q_y     & 1       & \dots & & \\
\vdots      & \vdots      & \vdots  & \dots & & \\
^j_{n_j}q_x & ^j_{n_j}q_y & 1       & \dots & & \\
            &             &         & ^j_1q_x     & ^j_1q_y     & 1      \\
            &             &         & \vdots      & \vdots      & \vdots \\
            &             &         & ^j_{n_j}q_x & ^j_{n_j}q_y & 1
\end{bmatrix},~
^j\SubMat{P}_\mathrm{c} =
\begin{bmatrix}
^j_1 p_x \\
\vdots \\
^j_{n_j} p_x \\
^j_1 p_y \\
\vdots \\
^j_{n_j} p_y
\end{bmatrix},
\end{equation*}
\normalsize%
in which $[^j_{n_j}p_x, ^j_{n_j}p_y]$ and $[^j_{n_j}q_x,  ^j_{n_j}q_y]$ are two
$x, y$ coordinates corresponding to images 1 and 2, in a coordinate system
local to each image.  It is instructive to keep the above non-reduced system in
mind, even though it does not have a solution.  To solve it, one image tile (in
the above case we choose tile 1) must be fixed.  This sets the frame of
reference for all other transformations, such that
\begin{equation}
\label{eq:2}%
\Mat{A}\Vec{x} - \Vec{b} =
\begin{bmatrix}
    ^{1,2}\SubMat{Q} & 0 \\
    0                & -^{1,3}\SubMat{Q} \\
    ^{2,3}\SubMat{P} & -^{2,3}\SubMat{Q}
\end{bmatrix}
\Vec{x} -
\begin{bmatrix}
    ^{1,2}\SubMat{P}_\textrm{c} \\
    ^{1,3}\SubMat{P}_\textrm{c} \\
    0
\end{bmatrix}.
\end{equation}
To fix a tile (done only for demonstration purposes here), the column range
corresponding to its transformation parameters is eliminated.  Vector $\Vec{b}$
is modified as shown in \cref{eq:2}.

For small numbers of tiles, solving \cref{eq:2} will yield the required
transformations using standard linear solvers (see \cref{sfig:1} for an
example with six tiles).

\subsection{Regularization}

\begin{figure}
\includegraphics[width=\textwidth]{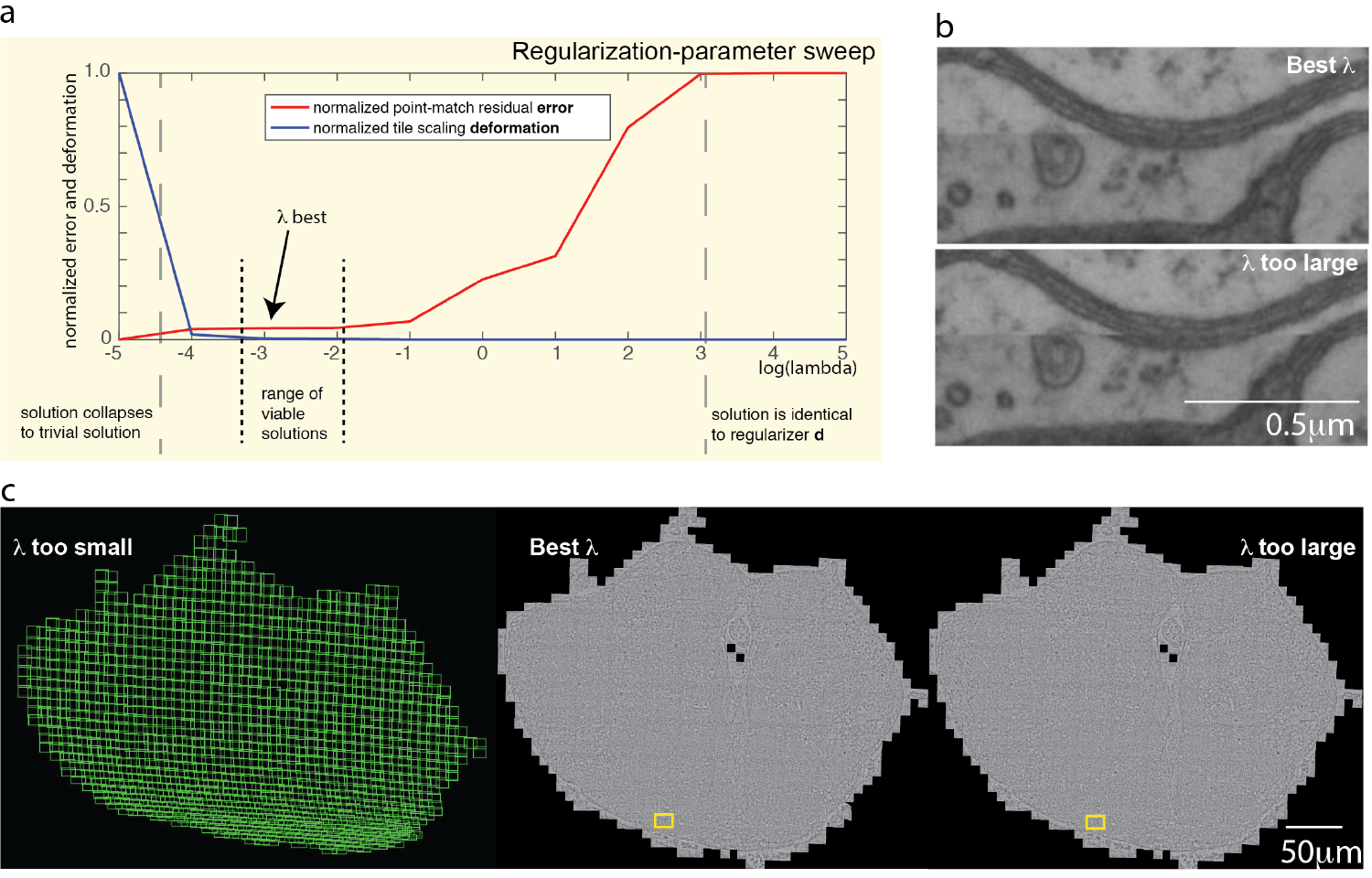}
\caption{Choice of regularization parameter. \textbf{(a--b)} Parameter sweep
(changing $\lambda$) for determining optimal range for regularization parameter
value. \textbf{(c)} Examples of solutions for three cases of regularization
parameter values ($\lambda$ too small, good, and too large) for a whole adult
fruit fly section. \textbf{(d)} and \textbf{(e)} Visual comparison of stitching
quality for cases between good regularization and excessive regularization.
Images correspond to approximately same region (boxes above).}
\label{fig:2}
\end{figure}

The strategy of fixing one tile is not sufficient in the general case.  With
increasing numbers of tiles, and when tiles are further removed from the fixed
reference tile, distortions are observed at the scale of the whole layer being
stitched (\cref{sfig:2,fig:2}c left panel).  To solve this problem, we include
explicit constraints, by adding a regularization term (term 2 in \cref{eq:1}).

Stage-reported tile coordinates may be accurate enough to be used as an
explicit constraint in special cases.  However, in the general case, these
numbers are not reliable.  We now describe a strategy to obtain a rough guess
($\Vec{d}$ in term 2 in \cref{eq:1}) to serve as regularizer. $^j_i{}p$
and $^j_i{}q$ are translated to their respective centers of mass, producing
$^j_i\hat{p}$ and $^j_i\hat{q}$ . A two dimensional affine transformation can
be constrained to a similarity deformation by considering, in addition to the
point-match data $^j_i\hat{p}$ and $^j_i\hat{q}$, the point-match data
subjected to an operator that transforms $(x,y)$ into $(-y,x)$
\citep{SchaeferAl06}.  We generalize this idea to the
full set of point-matches among all tiles.  The system is thereby implicitly
similarity-deformation-constrained by the data; both original and artificially
generated.  To accomplish this, we write an equation similar to Equation 2
for the joint system
\begin{equation}
\label{eq:3}%
\Mat{D}\Vec{m} - \Vec{f} =
\begin{bmatrix}
-^{1,2}\hat{\SubMat{Q}} & 0 \\
0                      & -^{1,3}\hat{\SubMat{Q}} \\
^{2,3}\hat{\SubMat{P}}  & -^{2,3}\hat{\SubMat{Q}}
\end{bmatrix}
\Vec{m} -
\begin{bmatrix}
^{1,2}\hat{\SubMat{P}}_\textrm{c} \\
^{1,3}\hat{\SubMat{P}}_\textrm{c} \\
0
\end{bmatrix},
\end{equation}
where $\Vec{m}$ is a column vector of coefficients (missing translation
terms).  The key difference is that blocks of \Mat{D} and \Vec{f} are now given
by
\scriptsize%
\begin{equation*}
^j\hat{\Mat{P}} =
\begin{bmatrix}
^j_1\hat{p}_x     & ^j_1\hat{p}_y      & & \\
\vdots            & \vdots             & & \\
^j_{n_j}\hat{p}_x & ^j_{n_j}\hat{p}_y  & & \\
^j_1\hat{p}_y     & -^j_1\hat{p}_x     & & \\
\vdots            & \vdots             & & \\
^j_{n_j}\hat{p}_y & -^j_{n_j}\hat{p}_x & & \\
& & ^j_1\hat{p}_x     & ^j_1\hat{p}_y  \\
& & \vdots            & \vdots \\
& & ^j_{n_j}\hat{p}_x & ^j_{n_j}\hat{p}_y \\
& & ^j_1\hat{p}_y     & -^j_1\hat{p}_x \\
& & \vdots            & \vdots \\
& & ^j_{n_j}\hat{p}_y & -^j_{n_j}\hat{p}_x
\end{bmatrix},~
^j\hat{\Mat{Q}} =
\begin{bmatrix}
^j_1\hat{q}_x     & ^j_1\hat{q}_y      & & \\
\vdots            & \vdots             & & \\
^j_{n_j}\hat{q}_x & ^j_{n_j}\hat{q}_y  & & \\
^j_1\hat{q}_y     & -^j_1\hat{q}_x     & & \\
\vdots            & \vdots             & & \\
^j_{n_j}\hat{q}_y & -^j_{n_j}\hat{q}_x & & \\
& & ^j_1\hat{q}_x     & ^j_1\hat{q}_y  \\
& & \vdots            & \vdots \\
& & ^j_{n_j}\hat{q}_x & ^j_{n_j}\hat{q}_y \\
& & ^j_1\hat{q}_y     & -^j_1\hat{q}_x \\
& & \vdots            & \vdots \\
& & ^j_{n_j}\hat{q}_y & -^j_{n_j}\hat{q}_x
\end{bmatrix},~
^j\hat{\Mat{P}}_\textrm{c} =
\begin{bmatrix}
^j_1\hat{p}_x \\
\vdots
^j_{n_j}\hat{p}_x \\
^j_1\hat{p}_y \\
\vdots
^j_{n_j}\hat{p}_y \\
^j_1\hat{p}_y \\
\vdots
^j_{n_j}\hat{p}_y \\
-^j_1\hat{p}_x \\
\vdots
-^j_{n_j}\hat{p}_x
\end{bmatrix}.
\end{equation*}
\normalsize%
When restricted to two images, this resembles the problem constructed by
\citet{SchaeferAl06}.

We solve \cref{eq:3} to obtain values \Vec{m} for all
transformation coefficients.  Due to the requirement of fixing one tile as
reference to be able to solve the matrix system, tiles that are far away from
the reference tile suffer excessive reduction in scale.  This is because in the limit of infinite overlapping tiles, and the fact that our deformation model is never fully accurate, error accumulates to a degree that drives the optimization to reduce this error by reducing overall scale of the tile collection.Therefore, all tiles
must be subsequently rescaled to their original area to yield the desired rotation
approximation $m_i$.  To obtain translation parameters $\Vec{t}$, we solve a
translation-only least-squares system separately.  The combined parameters
$m_i$  and $t_i$ are used to populate column vector $\Vec{d}$ in \cref{eq:1}.
So we write term 2 of \cref{eq:1} as
\begin{equation}
\label{eq:4}%
\Mat{B}\Vec{x} - \Vec{d} =
\begin{bmatrix}
\beta_1 & & & & & & \\
& \beta_2 & & & & & \\
& & \beta_3 & & & & \\
& & & \beta_4 & & & \\
& & & & \beta_5 & & \\
& & & & & \beta_6 & \\
& & & & & & \ddots
\end{bmatrix}
\Vec{x} -
\begin{bmatrix}
m_1 \\
m_2 \\
t_1 \\
m_3 \\
m_4 \\
t_2 \\
\vdots
\end{bmatrix},
\end{equation}

where \Mat{B} has dimensions $n \times n$, with diagonal elements determining
relative importance of regularization for a specific parameter.  \Mat{B} is the
identity matrix in most applications.  \Vec{d} is a column vector of length
$n$ and represents the approximate solution to the rigid-model problem.

Using \cref{eq:2,eq:4}, and choosing a suitable value for the regularization
term $\lambda$, we solve the full regularized system (\cref{eq:1}).  In normal
equation form
\begin{equation*}
\Vec{x} =
(\Mat{A}^\trnsp \cdot \Mat{D} \cdot \Mat{A} + \lambda\Mat{B}\trnsp
\cdot \Mat{B})^{-1} \cdot (\Mat{A}^\trnsp \cdot \Mat{D} \cdot \Vec{b} +
\lambda\Mat{B}^\trnsp \cdot \Vec{d}),
\end{equation*}
and assuming vector $\Vec{b}$ to generally be all zeros, \Mat{B} the identity
matrix and $\lambda$ a diagonal matrix (in the general case, each parameter
can be constrained independently)
\begin{equation*}
\Vec{x} =
(\Mat{A}^\trnsp \cdot \Mat{D} \cdot \Mat{A} + \lambda\Mat{I})^{-1} \cdot
(\lambda\Mat{B}^\trnsp \cdot \Vec{d}),
\end{equation*}
so we are solving a system
\begin{equation}
\label{eq:5}%
\tilde{\Mat{A}}\Vec{x} = \tilde{\Vec{b}} = 0, ~\text{where}~ \tilde{\Mat{A}} =
\Mat{A}^\trnsp \cdot \Mat{D} \cdot \Mat{A} + \lambda\Mat{I}, ~\text{and}~
\tilde{\Vec{b}} = \lambda \cdot \Vec{d}.
\end{equation}

\subsection{Choice of regularization parameters}

For problem sizes at the scale of several thousand tiles, solution of
\cref{eq:5} is fast ($< 1$\,s) and a parameter sweep for determining $\lambda$
is practical.  We calculate tile transformations and plot log $\lambda$
vs. a measure of deformation, which is taken to be the ratio of average area of
deformed tile relative to area of the undeformed tile (\cref{fig:2}a).

\subsection{Generalization to nonlinear transformations}

The matrix expressions above extend to nonlinear models in a straightforward
manner solely by modification of matrix blocks $^j\SubMat{P}$ and
$^j\SubMat{Q}$.

\subsection{Explicit constraints}

\Cref{eq:5} includes the regularization parameter $\lambda$, which in the
general case is a diagonal matrix with diagonal elements corresponding to the
degree of regularization desired for each individual parameter.  For example,
the user might want to decrease constraints on translation parameters of each
tile (leave $x$ and $y$ relatively free), and strongly constrain all parameters
(including translation) for one or more sections that should not be modified
by the solution.  Such strategies come in handy when performing local
improvements on alignment of especially problematic sections, while preventing
any perturbation of neighboring sections.

\section{Results}

We performed a series of registrations of ssTEM image data with increasing
numbers of tiles for both single section slices (montage registration) and
multiple-section volumes.

In all cases, tile metadata was first ingested into a dedicated
database.\footnote{Renderer: \href{https://github.com/saalfeldlab/render}{
https://github.com/saalfeldlab/render}}
Point-matches between potential tile-pairs were then calculated as by
\citet{SaalfeldAl12} and subsequently ingested into the database for retrieval
by the solver process to build a linear system (\cref{eq:5}).  Different linear
solvers were used to solve \cref{eq:5}.  All experiments were conducted on a 32
CPU Broadwell computer with 256\,GB RAM using Matlab version 2017a.  A parallel
pool with all 32 CPUs was used with parfor (parallel for loops) for constructing
the linear system.  An explicit parallel solution in Matlab was not used.  In
the case of PaStiX \citep{A:LaBRI::HRR01a}, we used a setup of 8 CPUs on a dedicated
Broadwell node.  Results are summarized in \cref{tab:1,tab:2}.  The purpose here
is to provide a general idea of CPU performance obtainable for such systems with
current hardware, and to compare direct vs. iterative linear solvers.

We observe that direct methods outperform iterative ones both in CPU time
requirements, per-tile point-match residual error and linear system precision
as expected.  For large systems with more than 1M tiles, PaStiX outperformed
other approaches significantly.  PaStiX is the only massively parallel linear
solver that we tested.  It is likely that other parallel direct solvers are
also suitable for this type of problem.

The efficiency of generating montage solutions makes a regularization
parameter-sweep practical.  \Cref{eq:5} was solved for a range of
$\lambda$ values for a 536-tile section dataset (\cref{fig:2}).  A
tile-deformation measure was determined as the mean deviation of tile areas
post-registration from starting undeformed tile areas.

\section{Discussion}

The presented method enables joint deformable registration of millions of images
using known linear algebra techniques.  Least-squares systems resembling
\cref{eq:1} are known in the literature \citep{GolubL13}.  The main contribution
of this work is enabling linear algebra solvers for the large EM
registration problem by providing an explicit matrix-based formulation for joint
estimation of a rigid-model approximation.  Without such a model, it is not
possible to use \cref{eq:1} for any but the most trivial problems.  Importantly,
parallel direct solvers can be used.

If the problem size (or hardware restriction) necessitates the use of
iterative methods over direct ones, then established iterative strategies such
as GMRES or stabilized biconjugate gradients may be used.  In this way, image
registration efforts for EM are decoupled from solver strategies and
automatically benefit from existing general scalable linear solvers and future
work to improve them.

\subsection{Code}

The computer code accompanying this work estimates transformation parameters
for translation, rigid approximation, affine and higher order polynomials up
to third degree.  It is written in the Matlab (The Mathworks Inc.) programming
language.  The main solver functions are summarized in \cref{tab:3} and
corresponding code can be obtained freely \citep{Khairy2018}.

\section{Acknowledgements}

We thank Davi Bock, Zhihao Zheng, Camenzind Robinson, Eric Perlman, Rick
Fetter and Nirmala Iyer for generating and providing us with the full adult fly
brain (FAFB) dataset.  We thank Bill Karsh and Lou Scheffer for discussions
about solver strategies.  We thank Eric Trautman, Tom Dolafi, Philipp
Hanslovsky, John Bogovic, Eric Perlman, Cristian Goina and Tom Kazimiers for
building computational tools to enable data management, retrieval, z-correction
and inspection.  We thank Goran Ceric, Rob Lines and Ken Carlile for help with
enabling PaSTiX computations on Janelia’s compute cluster.

\bibliography{references}

\begin{thebibliography}{17}
\providecommand{\natexlab}[1]{#1}
\providecommand{\url}[1]{\texttt{#1}}
\expandafter\ifx\csname urlstyle\endcsname\relax
  \providecommand{\doi}[1]{doi: #1}\else
  \providecommand{\doi}{doi: \begingroup \urlstyle{rm}\Url}\fi

\bibitem[Bock et~al.(2011)Bock, Lee, Kerlin, Andermann, Hood, Wetzel,
  Yurgenson, Soucy, Kim, and Reid]{BockAl11}
D.~D. Bock, W.-C.~A. Lee, A.~M. Kerlin, M.~L. Andermann, G.~Hood, A.~W. Wetzel,
  S.~Yurgenson, E.~R. Soucy, H.~S. Kim, and R.~C. Reid.
\newblock Network anatomy and in vivo physiology of visual cortical neurons.
\newblock \emph{Nature}, 471:\penalty0 177--182, 2011.
\newblock \doi{doi:10.1038/nature09802}.

\bibitem[Bookstein(1989)]{Bookstein89}
F.~L. Bookstein.
\newblock Principal warps: thin-plate splines and the decomposition of
  deformations.
\newblock \emph{IEEE Transactions on Pattern Analysis and Machine
  Intelligence}, 11\penalty0 (6):\penalty0 567--585, 1989.

\bibitem[Cardona et~al.(2012)Cardona, Saalfeld, Schindelin, Arganda-Carreras,
  Preibisch, Longair, Toman\v{c}\'{a}k, Hartenstein, and Douglas]{CardonaAl12}
A.~Cardona, S.~Saalfeld, J.~Schindelin, I.~Arganda-Carreras, S.~Preibisch,
  M.~Longair, P.~Toman\v{c}\'{a}k, V.~Hartenstein, and R.~J. Douglas.
\newblock Trakem2 software for neural circuit reconstruction.
\newblock \emph{PLoS ONE}, 7\penalty0 (6):\penalty0 e38011, 2012.
\newblock \doi{10.1371/journal.pone.0038011}.

\bibitem[Golub and Loan(2013)]{GolubL13}
G.~Golub and C.~V. Loan.
\newblock \emph{Matrix Computations}.
\newblock The Johns Hopkins University Press, 2013.

\bibitem[H\'enon et~al.(2002)H\'enon, Ramet, and Roman]{A:LaBRI::HRR01a}
P.~H\'enon, P.~Ramet, and J.~Roman.
\newblock {PaStiX}: {A} {H}igh-{P}erformance {P}arallel {D}irect {S}olver for
  {S}parse {S}ymmetric {D}efinite {S}ystems.
\newblock \emph{{P}arallel {C}omputing}, 28\penalty0 (2):\penalty0 301--321,
  Jan. 2002.

\bibitem[Karsh(2016)]{Karsh2016}
B.~Karsh.
\newblock Aligner for large scale serial section image data.
\newblock \emph{GitHub repository},
  \url{https://github.com/billkarsh/Alignment_Projects}, 2016.

\bibitem[Kaynig et~al.(2010)Kaynig, Fischer, M{\"u}ller, and
  Buhmann]{KaynigAl10a}
V.~Kaynig, B.~Fischer, E.~M{\"u}ller, and J.~M. Buhmann.
\newblock Fully automatic stitching and distortion correction of transmission
  electron microscope images.
\newblock \emph{Journal of Structural Biology}, 171\penalty0 (2):\penalty0
  163--173, 2010.

\bibitem[Khairy(2018)]{Khairy2018}
K.~A. Khairy.
\newblock {EM} aligner.
\newblock \emph{GitHub repository},
  \url{https://github.com/khaledkhairy/EM_aligner}, 2018.

\bibitem[Lowe(2004)]{Lowe04}
D.~G. Lowe.
\newblock Distinctive image features from scale-invariant keypoints.
\newblock \emph{International Journal of Computer Vision}, 60\penalty0
  (2):\penalty0 91--110, 2004.

\bibitem[Saalfeld et~al.(2010)Saalfeld, Cardona, Hartenstein, and
  Toman\v{c}\'{a}k]{SaalfeldAl10}
S.~Saalfeld, A.~Cardona, V.~Hartenstein, and P.~Toman\v{c}\'{a}k.
\newblock As-rigid-as-possible mosaicking and serial section registration of
  large sstem datasets.
\newblock \emph{Bioinformatics}, 26\penalty0 (12):\penalty0 i57--i63, 2010.
\newblock \doi{10.1093/bioinformatics/btq219}.

\bibitem[Saalfeld et~al.(2012)Saalfeld, Fetter, Cardona, and
  Toman\v{c}\'{a}k]{SaalfeldAl12}
S.~Saalfeld, R.~Fetter, A.~Cardona, and P.~Toman\v{c}\'{a}k.
\newblock Elastic volume reconstruction from series of ultra-thin microscopy
  sections.
\newblock \emph{Nature Methods}, 9\penalty0 (7):\penalty0 717--720, 2012.
\newblock \doi{10.1038/nmeth.2072}.

\bibitem[Schaefer et~al.(2006)Schaefer, McPhail, and Warren]{SchaeferAl06}
S.~Schaefer, T.~McPhail, and J.~Warren.
\newblock Image deformation using moving least squares.
\newblock \emph{ACM Transactions on Graphics}, 25\penalty0 (3):\penalty0
  533--540, 2006.
\newblock \doi{http://doi.acm.org/10.1145/1141911.1141920}.

\bibitem[Scheffer et~al.(2013)Scheffer, Karsh, and Vitaladevun]{SchefferAl13}
L.~K. Scheffer, B.~Karsh, and S.~Vitaladevun.
\newblock Automated alignment of imperfect em images for neural reconstruction.
\newblock \emph{arXiv:1304.6034 [q-bio.QM]}, 2013.

\bibitem[Takemura et~al.(2015)Takemura, Xu, Lu, Rivlin, Parag, Olbris, Plaza,
  Zhao, Katz, Umayam, Weaver, Hess, Horne, Nunez-Iglesias, Aniceto, Chang,
  Lauchie, Nasca, Ogundeyi, Sigmund, Takemura, Tran, Langille, Le~Lacheur,
  McLin, Shinomiya, Chklovskii, Meinertzhagen, and Scheffer]{TakemuraAl2015}
S.-y. Takemura, C.~S. Xu, Z.~Lu, P.~K. Rivlin, T.~Parag, D.~J. Olbris,
  S.~Plaza, T.~Zhao, W.~T. Katz, L.~Umayam, C.~Weaver, H.~F. Hess, J.~A. Horne,
  J.~Nunez-Iglesias, R.~Aniceto, L.-A. Chang, S.~Lauchie, A.~Nasca,
  O.~Ogundeyi, C.~Sigmund, S.~Takemura, J.~Tran, C.~Langille, K.~Le~Lacheur,
  S.~McLin, A.~Shinomiya, D.~B. Chklovskii, I.~A. Meinertzhagen, and L.~K.
  Scheffer.
\newblock Synaptic circuits and their variations within different columns in
  the visual system of drosophila.
\newblock \emph{Proceedings of the National Academy of Sciences}, 112\penalty0
  (44):\penalty0 13711--13716, 2015.
\newblock ISSN 0027-8424.
\newblock \doi{10.1073/pnas.1509820112}.

\bibitem[Tasdizen et~al.(2010)Tasdizen, Koshevoy, Grimm, Anderson, Jones, Watt,
  Whitaker, and Marc]{TasdizenAl10}
T.~Tasdizen, P.~Koshevoy, B.~C. Grimm, J.~R. Anderson, B.~W. Jones, C.~B. Watt,
  R.~T. Whitaker, and R.~E. Marc.
\newblock Automatic mosaicking and volume assembly for high-throughput
  serial-section transmission electron microscopy.
\newblock \emph{Journal of Neuroscience Methods}, 193\penalty0 (1):\penalty0
  132--144, 2010.
\newblock \doi{10.1016/j.jneumeth.2010.08.001}.

\bibitem[ya~Takemura et~al.(2013)ya~Takemura, Bharioke, Lu, Nern, Vitaladevuni,
  Rivlin, Katz, Olbris, Plaza, Winston, Zhao, Horne, Fetter, Takemura, Blazek,
  Chang, Ogundeyi, Saunders, Shapiro, Sigmund, Rubin, Scheffer, Meinertzhagen,
  and Chklovskii]{TakemuraAl2013}
S.~ya~Takemura, A.~Bharioke, Z.~Lu, A.~Nern, S.~Vitaladevuni, P.~K. Rivlin,
  W.~T. Katz, D.~J. Olbris, S.~M. Plaza, P.~Winston, T.~Zhao, J.~A. Horne,
  R.~D. Fetter, S.~Takemura, K.~Blazek, L.-A. Chang, O.~Ogundeyi, M.~A.
  Saunders, V.~Shapiro, C.~Sigmund, G.~M. Rubin, L.~K. Scheffer, I.~A.
  Meinertzhagen, and D.~B. Chklovskii.
\newblock A visual motion detection circuit suggested by {Drosophila}
  connectomics.
\newblock \emph{Nature}, 500:\penalty0 175--181, 2013.
\newblock \doi{10.1038/nature12450}.

\bibitem[Zheng et~al.(2017)Zheng, Lauritzen, Perlman, Robinson, Nichols,
  Milkie, Torrens, Price, Fisher, Sharifi, Calle-Schuler, Kmecova, Ali, Karsh,
  Trautman, Bogovic, Hanslovsky, Jefferis, Kazhdan, Khairy, Saalfeld, Fetter,
  and Bock]{ZhengAl2017}
Z.~Zheng, J.~S. Lauritzen, E.~Perlman, C.~G. Robinson, M.~Nichols, D.~Milkie,
  O.~Torrens, J.~Price, C.~B. Fisher, N.~Sharifi, S.~A. Calle-Schuler,
  L.~Kmecova, I.~J. Ali, B.~Karsh, E.~T. Trautman, J.~Bogovic, P.~Hanslovsky,
  G.~S. X.~E. Jefferis, M.~Kazhdan, K.~Khairy, S.~Saalfeld, R.~D. Fetter, and
  D.~D. Bock.
\newblock A complete electron microscopy volume of the brain of adult
  drosophila melanogaster.
\newblock \emph{bioRxiv}, page 140905, 2017.

\end{thebibliography}

\begin{landscape}
\thispagestyle{empty}
\begin{table}[h]
\footnotesize
\begin{tabularx}{\linewidth}%
{@{}>{\hsize=1.25\hsize}X>{\hsize=.5\hsize}XX>{\hsize=1.25\hsize}XXXXXXX@{}}
Section ID &
\#tiles &
Solver Method &
Solver strategy &
solve $\Mat{A}\times\Vec{b}$ time [s] &
Mean residual error per tile [px] &
Precision* &
non-zeros \Mat{A} &
matrix \Mat{A} generation time [s] &
\#point-matches \\ \hline
200 (FAFB) &
158 &
MB &
direct &
0.001244 &
0.16122 &
6.96E-12 &
21840 &
0.2156 &
3640 \\
2000 (FAFB) &
1824 &
MB &
direct &
0.009751 &
0.11166 &
6.86E-12 &
2.62E+05 &
0.34587 &
43620 \\
3000 (FAFB) &
2764 &
MB &
direct &
0.021364 &
0.18095 &
8.81E-12 &
4.80E+05 &
0.50232 &
79940 \\
4500 (FAFB) &
3100 &
MB &
direct &
0.022424 &
0.12806 &
7.02E-12 &
4.46E+05 &
0.46282 &
74380 \\
(Dataset2) &
6013 &
MB &
direct &
0.094284 &
0.4568 &
1.88E-11 &
1.84E+06 &
1.4129 &
3.06E+05 \\ \hline
200 (FAFB) &
158 &
PaStiX &
direct/parallel &
0.16033 &
0.16277 &
6.22E-12 &
21840 &
0.2183 &
3640 \\
2000 (FAFB) &
1824 &
PaStiX &
direct/parallel &
0.05585 &
0.11291 &
6.64E-12 &
2.62E+05 &
0.34861 &
43620 \\
3000 (FAFB) &
2764 &
PaStiX &
direct/parallel &
0.065824 &
0.18074 &
8.50E-12 &
4.80E+05 &
0.48846 &
79940 \\
4500 (FAFB) &
3100 &
PaStiX &
direct/parallel &
0.09502 &
0.12727 &
6.78E-12 &
4.46E+05 &
0.49057 &
74380 \\
(Dataset2) &
6013 &
PaStiX &
direct/parallel &
0.16285 &
0.45634 &
1.73E-11 &
1.84E+06 &
1.2931 &
3.06E+05 \\ \hline
200 (FAFB) &
158 &
MBCG &
iterative &
1.0337 &
0.16263 &
4.00E-07 &
21840 &
0.22202 &
3640 \\
2000 (FAFB) &
1824 &
MBCG &
iterative &
7.8858 &
0.11412 &
1.34E-06 &
2.62E+05 &
0.35856 &
43620 \\
3000 (FAFB) &
2764 &
MBCG &
iterative &
11.082 &
0.18075 &
1.08E-07 &
4.80E+05 &
0.49591 &
79940 \\
4500 (FAFB) &
3100 &
MBCG &
iterative &
11.598 &
0.12971 &
1.80E-06 &
4.46E+05 &
0.4871 &
74380 \\
(Dataset2) &
6013 &
MBCG &
iterative &
28.979 &
0.46127 &
0.00033331 &
1.84E+06 &
1.2815 &
3.06E+05 \\ \hline
200 (FAFB) &
158 &
MGMRES &
iterative &
11.255 &
0.17344 &
2.57E-05 &
21840 &
0.20753 &
3640 \\
2000 (FAFB) &
1824 &
MGMRES &
iterative &
123.98 &
0.13031 &
8.56E-06 &
2.62E+05 &
0.34978 &
43620 \\
3000 (FAFB) &
2764 &
MGMRES &
iterative &
154.12 &
0.25182 &
2.44E-05 &
4.80E+05 &
0.50051 &
79940 \\
4500 (FAFB) &
3100 &
MGMRES &
iterative &
159.14 &
0.14295 &
5.39E-06 &
4.46E+05 &
0.57044 &
74380 \\
(Dataset2) &
6013 &
MGMRES &
iterative &
317.55 &
0.4889 &
0.0030296 &
1.84E+06 &
1.291 &
3.06E+05
\end{tabularx}
\vspace{1em}

MB = Matlab backslash operator \\
PaStiX = Parallel sparse matrix solver \\
MBCG = Matlab biconjugate gradient solver \\
MGMRES = Matlab GMRES solver

\normalsize
\caption{Comparison of solver performance for montage solutions with increasing
numbers of tiles, solver codes and solver strategies.}
\label{tab:1}
\end{table}

\begin{table}[h]
\footnotesize
\begin{tabularx}{\linewidth}%
{@{}>{\hsize=.75\hsize}XX>{\hsize=1.25\hsize}XXXXXXX@{}}
\#tiles &
Solver Method &
Solver strategy &
solve $\Mat{A}\times\Vec{b}$ time [s] &
Mean residual error per tile [px] &
Precision* &
non-zeros \Mat{A} &
matrix \Mat{A} generation time [s] &
\#point-matches \\ \hline
99775 &
MB &
direct &
11.13 &
3.11 &
0.86E-13 &
22147944 &
63.88 &
3691324 \\
499949 &
MB &
direct &
295.31 &
4.08 &
0.99E-13 &
138059052 &
397.39 &
23009842 \\
999514 &
MB &
direct &
1343.53 &
3.45 &
0.12E-13 &
316543020 &
1002.35 &
52757170 \\ \hline
99775 &
PaStiX &
direct/parallel &
2.81 &
3.10 &
1.88E-13 &
22146972 &
67.83 &
3691162 \\
499949 &
PaStiX &
direct/parallel &
24.29 &
4.07 &
2.80E-13 &
138074232 &
412.74 &
23012372 \\
999514 &
PaStiX &
direct/parallel &
80.18 &
3.46 &
3.89E-13 &
316523388 &
947.75 &
52753898 \\ \hline
99775 &
MBCG &
Iterative &
2116.64 &
3.11 &
0.84E-13 &
22149816 &
64.07 &
3691636
\end{tabularx}

\normalsize
\caption{Comparison of solver performance for massive data with increasing
numbers of tiles.}
\label{tab:2}
\end{table}

\begin{table}[h]
\footnotesize
\begin{tabularx}{\linewidth}%
{@{}>{\hsize=.5\hsize}X>{\hsize=.72\hsize}X>{\hsize=0.17\hsize}XX@{}}
Model &
EM\_aligner function name &
specify $\lambda$ &
Adjustable parameters \\ \hline
Translation &
system\_solve\_translation &
no &
Point-match filtering,\newline
Min. and max. number of point matches,\newline
Solver type (`backslash', `pastix', `bicgstab', `gmres') \\ \hline
Rigid approximation &
system\_solve\_rigid\_approximation &
no &
Same as above \\ \hline
Affine &
system\_solve &
yes &
Same as above, specify rough alignment \\
 &
system\_solve\_affine\_with\_constraints &
yes &
Same as above, specify section-wise constraints \\ \hline
Higher order polynomial &
system\_solve\_polynomial &
yes &
Same as Affine, specify polynomial degree
\end{tabularx}

\normalsize
\caption{Types of solutions available with EM\_aligner.}
\label{tab:3}
\end{table}
\end{landscape}

\begin{figure}
\includegraphics[width=\textwidth]{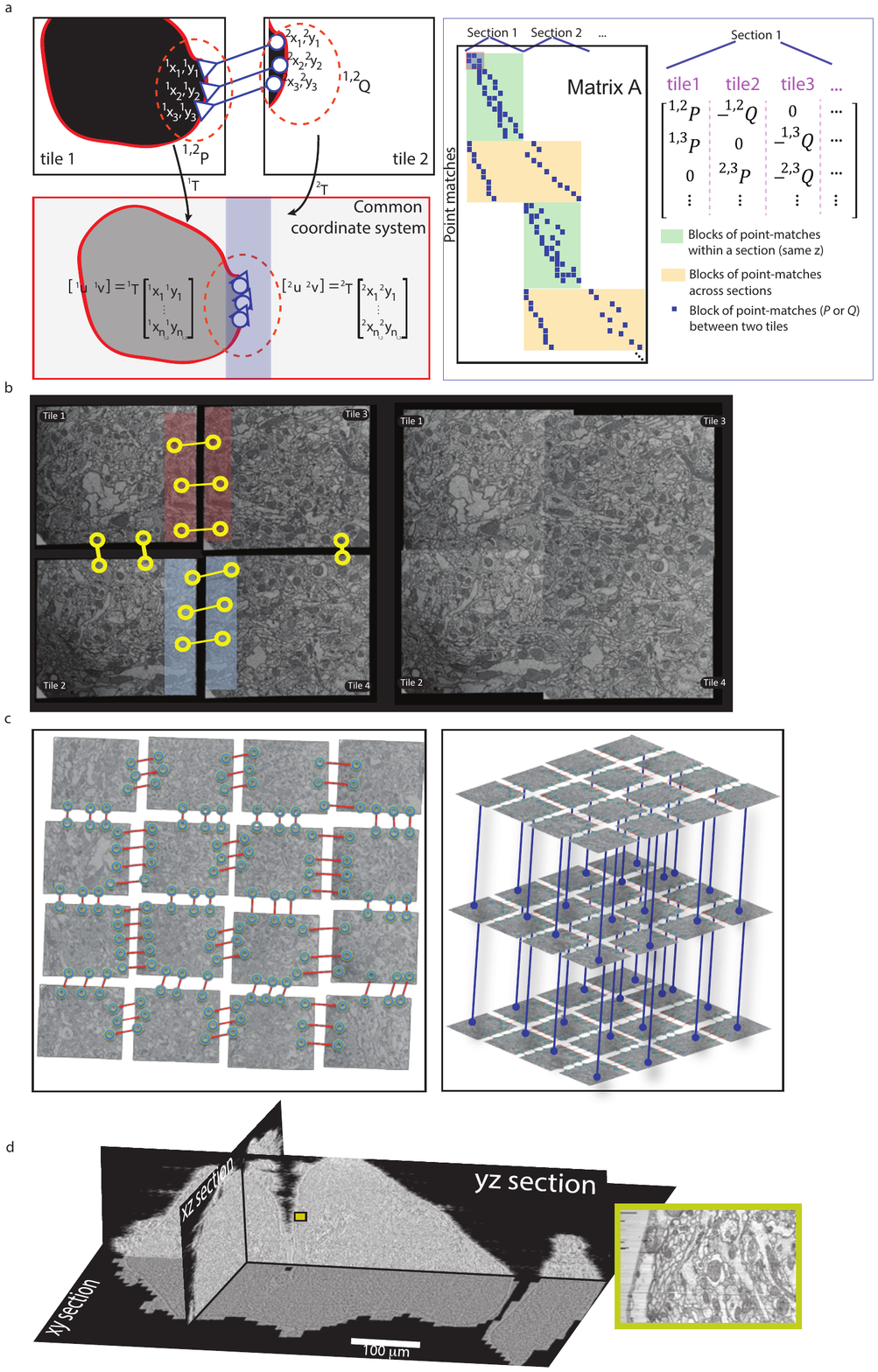}
\end{figure}
\begin{figure}
\caption{Stitching overlapping image tiles, that constitute a
contiguous three-
dimensional image volume, by minimizing point-match distances. \textbf{(a)}
(Upper left panels) Cartoon of two partially overlapping tiles with point match
sets that have been found by matching features.  Matching point-pairs
constitute correspondences in set $\SubMat{P}$ and $\SubMat{Q}$ for each tile
respectively.  (Lower left panel) Optimal transformations T1 and T2 when
applied to their respective tiles, transform them into a common coordinate
system $(u, v)$ in which they are registered seamlessly.  Right panel: Matrix
$\Mat{A}$ (of \cref{eq:1}) is populated by blocks of $\SubMat{P}$ and
$\SubMat{Q}$ corresponding to matching point sets. \textbf{(b)} Example of four
individual tiles with overlapping areas (shaded color) and a subset of point
matches (connected yellow circles), before and after stitching using
\cref{eq:2}.  \textbf{(c)} Typically, tiles are connected by large numbers of
sets of point-matches within (left panel) and across sections (right panel).
The full set of point-matches is used to construct a linear least-squares system
(\cref{eq:5}) the solution of which is an estimate of the joint deformations of
all tiles necessary to minimize all point-match distances simultaneously.
\textrm{(d)} Large-scale affine-model volume alignment of 2.67 million tiles
across $\approx$2500 ssTEM sections of the FAFB dataset.  The solver used was
PaStiX \citep{A:LaBRI::HRR01a}.  The system was regularized against rough-aligned data
using a rigid-approximation solution.  The rough alignment process aligns
images of section montages (montage-scapes) obtained using rigid approximation
followed by affine refinement.}
\label{fig:1}
\end{figure}

\begin{figure}
\includegraphics[width=\textwidth]{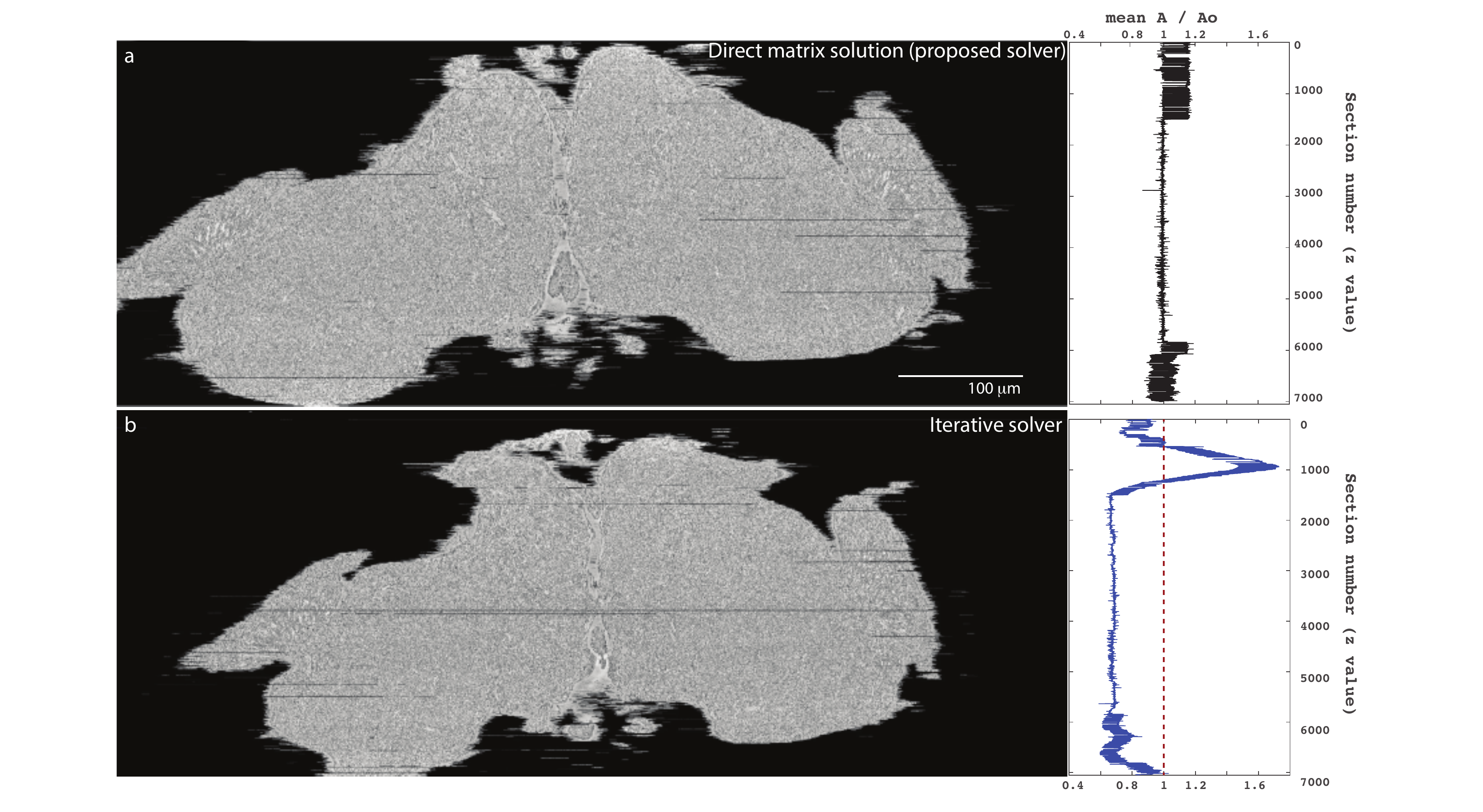}
\caption{Comparison of direct matrix solver method with iterative solution by
\citet{SchefferAl13}.  \textbf{(a)} The volume was subdivided into 78
contiguous overlapping slabs of approximately 250,000 tiles each.  Slab overlap
was 100 sections.  Each slab was solved using Matlab’s backslash operator.
Total matrix solution time $\approx 18$~hours using one Broadwell node with 32~CPUs; 576 CPU hours. Right panel shows deformation measure as ratio between tile area after optimization vs. original tile area for 7062 sections \textbf{(b)}
Full system solve by iteratively updating transformations for each tile, solving
a small affine system locally using the method in \citet{SchefferAl13}.  Total
solution time 6~hours on 240~CPUs; 1,440 CPU hours. Right panel as in \textbf{(a)}.}
\label{fig:3}
\end{figure}

\begin{figure}
\includegraphics[width=\textwidth]{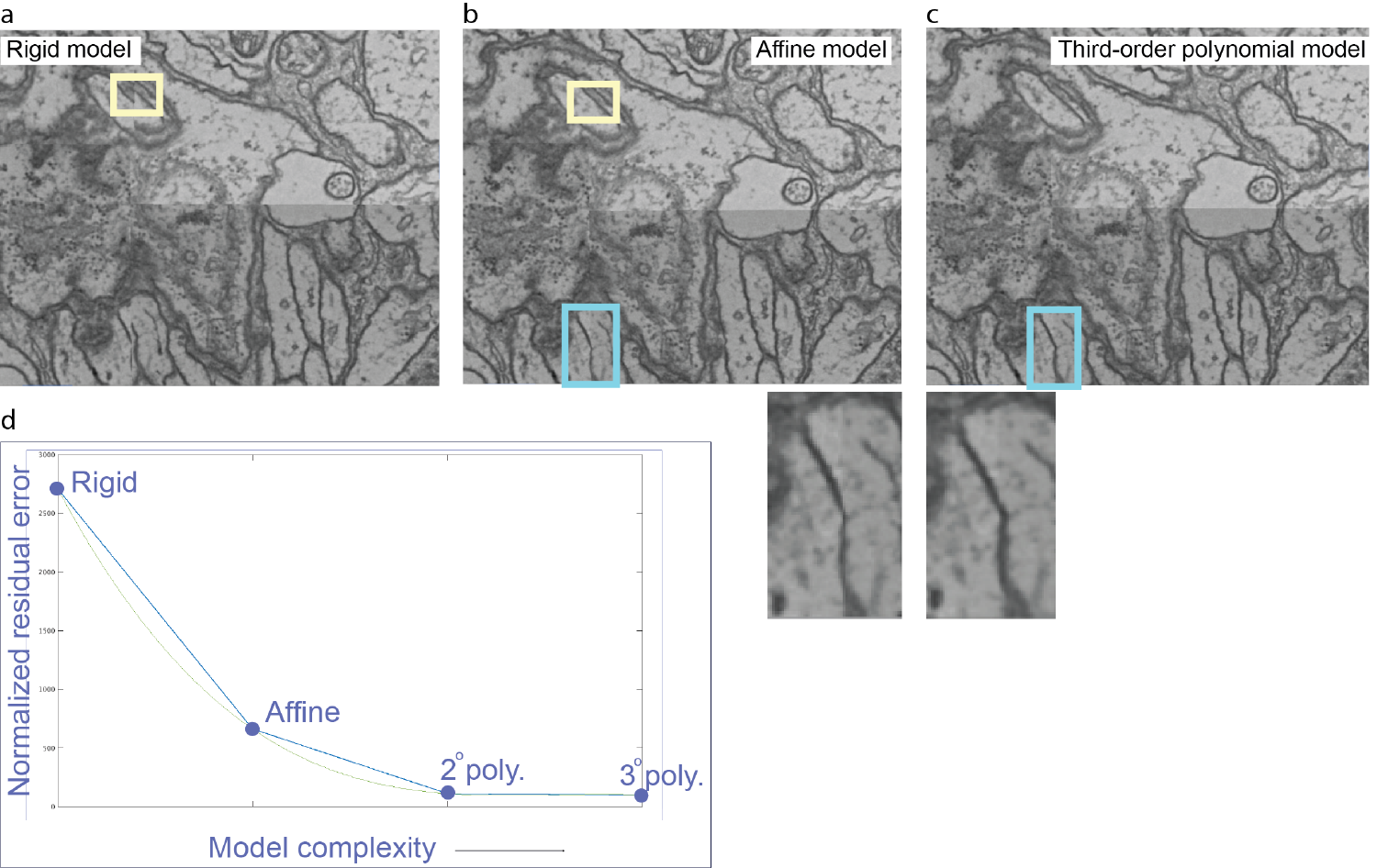}
\caption{Nonlinear models are supported by the proposed joint-deformation
solution method.}
\label{fig:4}
\end{figure}

\renewcommand{\figurename}{Supplementary Figure}
\setcounter{figure}{0}

\begin{figure}
\includegraphics[width=\textwidth]{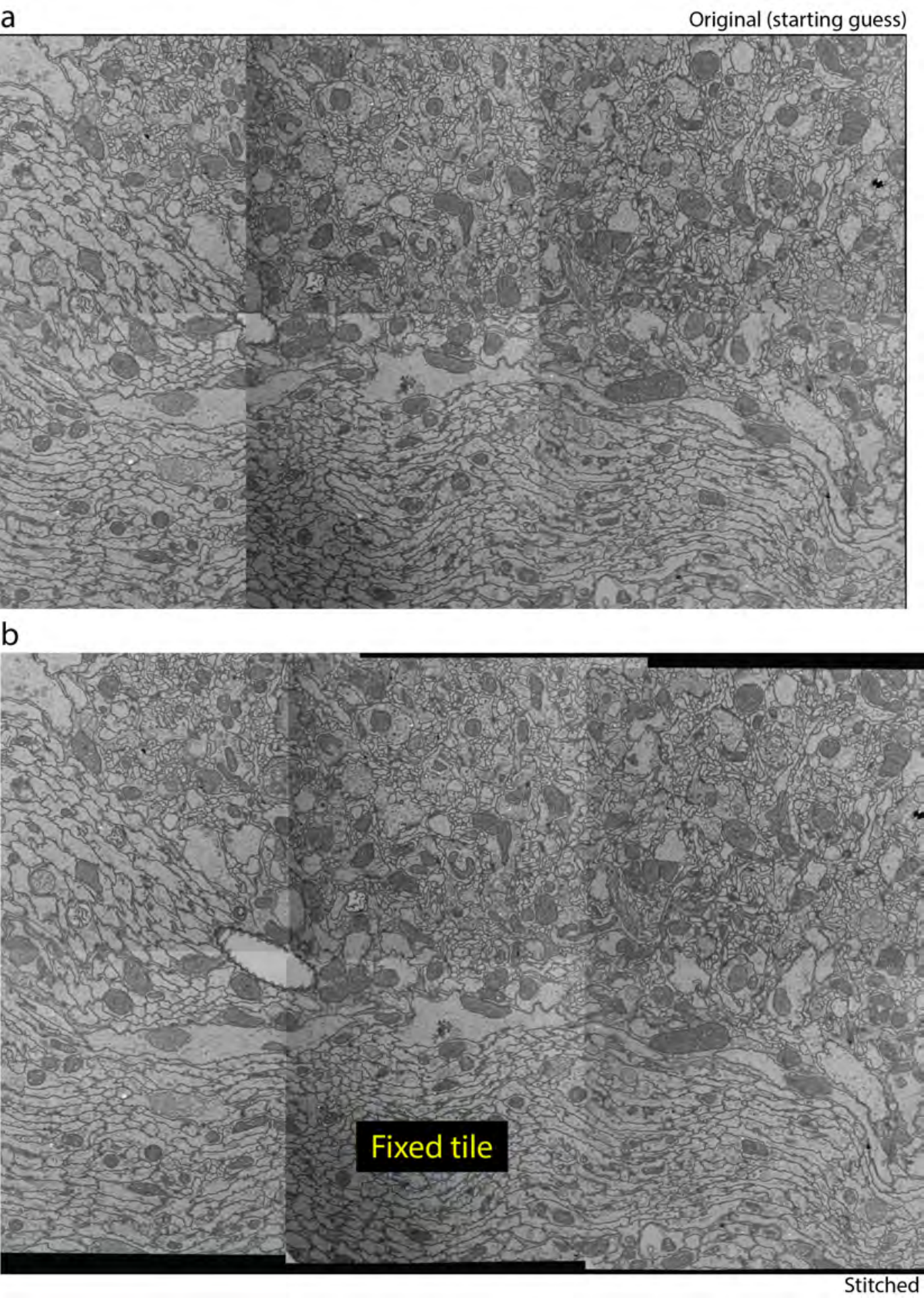}
\caption{Stitching of six tiles without regularization.  Only term 1 in
\cref{eq:1} was constructed and solved.  One tile, labeled ``Fixed tile'', was
used as reference.}
\label{sfig:1}
\end{figure}

\begin{figure}
\includegraphics[width=\textwidth]{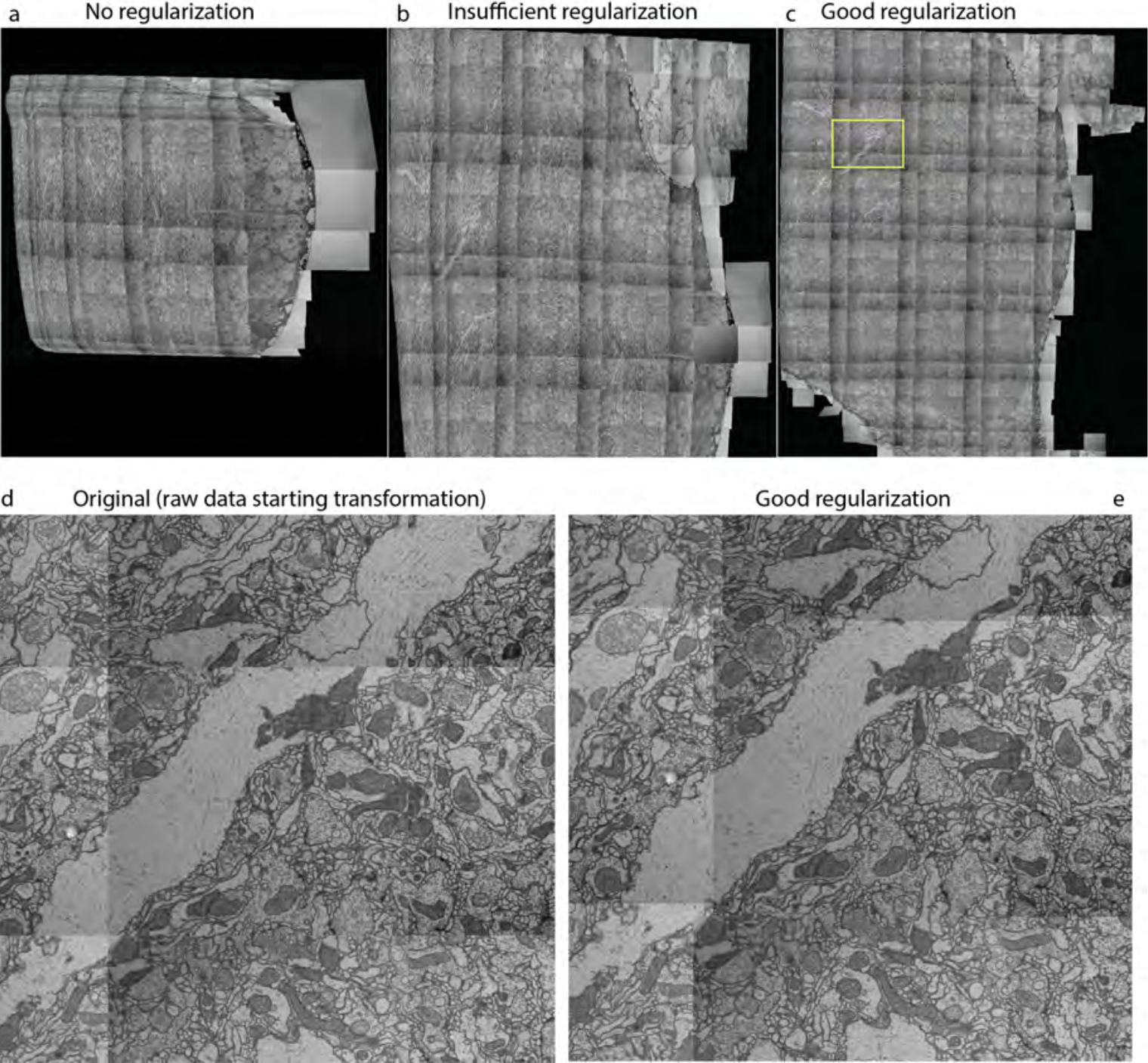}
\caption{Necessity of regularization for large numbers of tiles.
\textbf{(a--c)} 536~tiles of an adult fruit-fly section registered without
regularization ($\lambda = 0$), with insufficient, and with good regularization
respectively. \textbf{(d,e)} zoomed-in view of region marked by yellow box in
\textbf{(c)} for the initial guess (before stitching) and the regularized
stitching result respectively.}
\label{sfig:2}
\end{figure}

\begin{figure}
\includegraphics[width=0.75\textwidth]{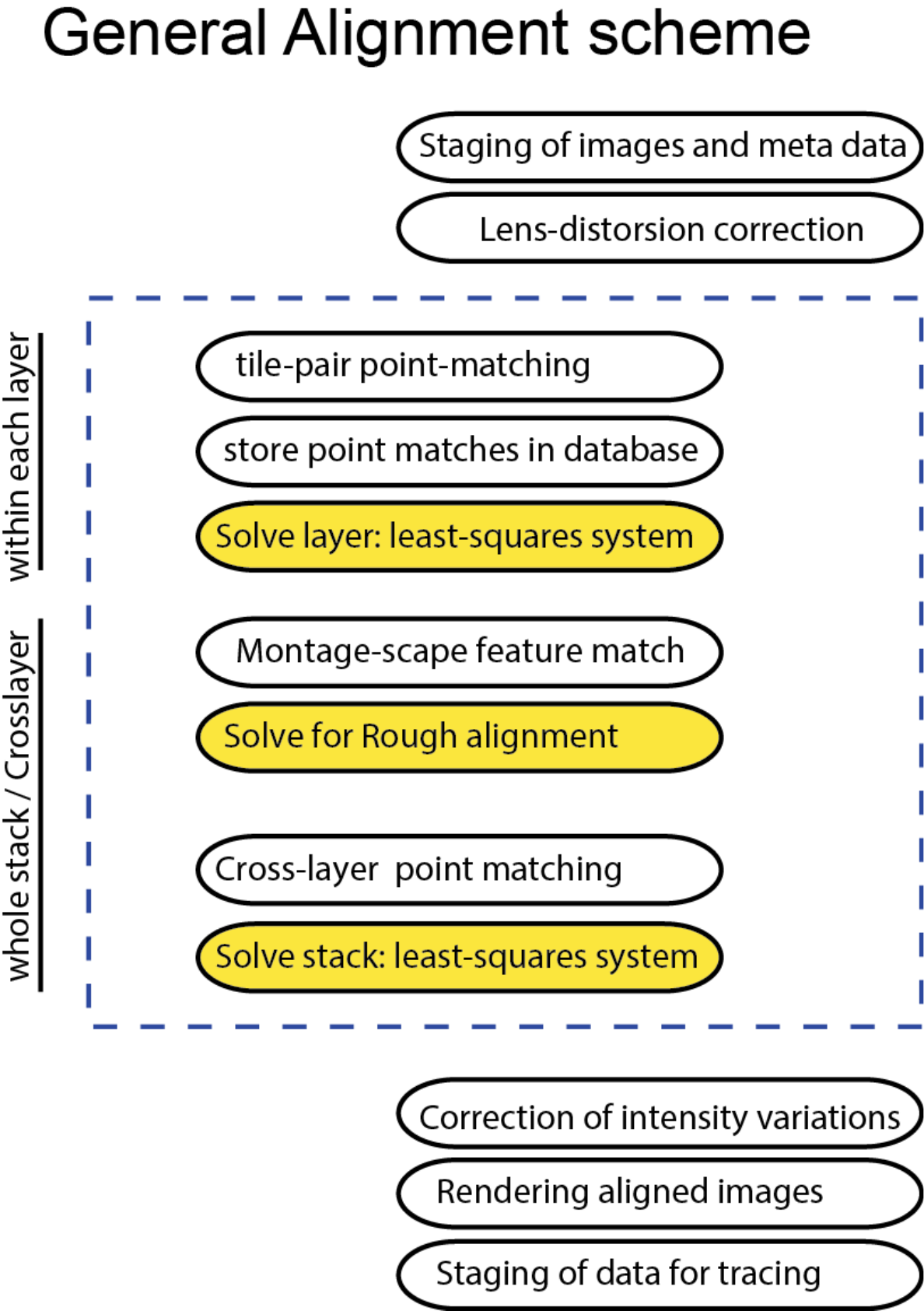}
\caption{Typical workflow for serial section montaging and alignment.}
\label{sfig:3}
\end{figure}

\end{document}